\documentclass{bmvc2k}

\usepackage{times}
\usepackage{epsfig}
\usepackage{graphicx}
\usepackage{amsmath}
\usepackage{amssymb}
\usepackage{bm}
\usepackage{diagbox}
\usepackage{multirow}
\usepackage{graphicx}
\usepackage{booktabs}
\usepackage{soul}
\usepackage[dvipsnames]{xcolor}
\usepackage{changes}
\usepackage{verbatim}
\usepackage{footnote}
\usepackage{wrapfig}
\makesavenoteenv{tabular}
\newcommand{\stkout}[1]{\ifmmode\text{\sout{\ensuremath{#1}}}\else\sout{#1}\fi}
\setdeletedmarkup{\stkout{#1}}

\newcommand{\head}[1]{\noindent\textbf{#1}}

\usepackage{mathtools}

\title{Dynamic Graph Modules for \\Modeling Object-Object Interactions in Activity Recognition}

\addauthor{Hao Huang}{hhuang40@ur.rochester.edu}{1}
\addauthor{Luowei Zhou}{luozhou@umich.edu}{2}
\addauthor{Wei Zhang}{wzhang45@ur.rochester.edu}{1}
\addauthor{Jason J. Corso}{jjcorso@umich.edu}{2}
\addauthor{Chenliang Xu}{chenliang.xu@rochester.edu}{1}

\addinstitution{
 University of Rochester\\
 Rochester, New York, USA
}
\addinstitution{
 University of Michigan,\\
 Ann Arbor, Michigan, USA
}

\runninghead{Huang, Zhou, Zhang, Corso, Xu}{Action Recognition \& Dynamic Graph}

\def\eg{\emph{e.g}\bmvaOneDot}

\def\etal{\emph{et al}\bmvaOneDot}

\begin{document}

\maketitle

\begin{abstract}

Video action recognition, a critical problem in video understanding, has been gaining increasing attention. To identify actions induced by complex object-object interactions, we need to consider not only spatial relations among objects in a single frame, but also temporal relations among different or the same objects across multiple frames. However, existing approaches that model video representations and non-local features are either incapable of explicitly modeling relations at the object-object level or unable to handle streaming videos. In this paper, we propose a novel dynamic hidden graph module to model complex object-object interactions in videos, of which two instantiations are considered: a visual graph that captures appearance/motion changes among objects and a location graph that captures relative spatiotemporal position changes among objects. Additionally, the proposed graph module allows us to process streaming videos, setting it apart from existing methods.
Experimental results on benchmark datasets, Something-Something and ActivityNet, show the competitive performance of our method.

\end{abstract}

\section{Introduction}
\label{sec:intro}

\begin{figure*}
\centering
\includegraphics[width=0.8\linewidth]{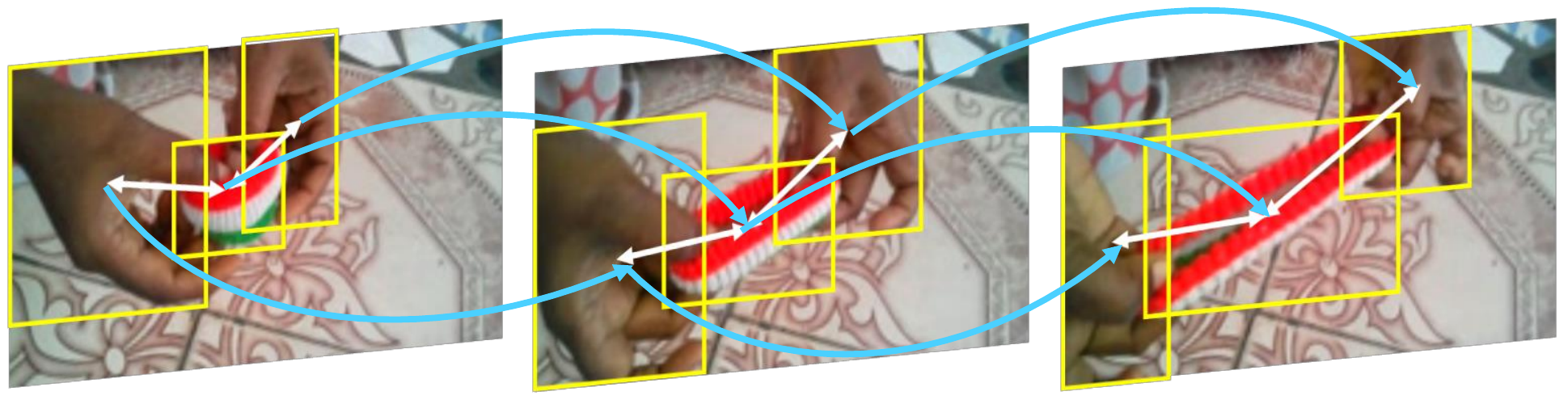}
  \caption{The action ``pulling two ends of a hair band so that it gets stretched'' contains interactions between two hands and a hair band. The visual graph captures the relation between visually similar objects (blue arrows) while the location graph captures relation between overlapped or close objects (white arrows).}
\label{fig:teaser}\vspace{-10pt}
\end{figure*}

Video action recognition has shown remarkable progress through the use of deep learning~\cite{karpathy2014large, simonyan2014two, tran2017convnet, carreira2017quo, wang2016temporal} and newly-released datasets, \eg, Kinetics~\cite{kay2017kinetics}, Something-Something~\cite{goyal2017something, mahdisoltani2018fine}, and ActivityNet~\cite{caba2015activitynet} to name a few. Despite the importance of complex object-object interactions in defining actions (see Fig.~\ref{fig:teaser} for an example), they are often overlooked. To recognize such interactions, we postulate that two general relations should be taken into consideration: 1) the interactions among different objects in a single frame, and 2) the transition of such interactions among different objects and the same object across multiple frames. We denote the former relation as spatial relation, and the latter one as temporal relation. Both are crucial to recognize actions involving multiple objects. An effective action recognition model should be able to capture both relations precisely and simultaneously. 


Despite many recent works~\cite{battaglia2016interaction, hu2018relation, watters2017visual, qi2018learning, wang2017appearance} that explore modeling interactions between objects, few of them build models to capture the spatiotemporal interactions simultaneously. To model interactions among objects in both the spatial and temporal domain, 
we propose a dynamic graph module to capture object interactions from the beginning of a video in a progressive way to recognize actions. Similar to LSTM, we maintain a hidden state across time steps, in the form of a complete directed graph with self-connections, which we named \textbf{hidden graph}. When a new frame arrives, regions of interests (RoIs) ~\cite{girshick2015fast,ren2015faster} in this frame are connected with nodes in the hidden graph by edges. Then, messages from RoIs in the new arriving frame will be passed to the hidden graph explicitly. After the information passing, the hidden graph further performs a self-update. 
A global aggregation function is applied to summarize the hidden graph for action recognition at this time step. When the next frame arrives, we repeat the above steps.
Through this dynamic hidden graph structure, we capture both the spatial relation in each arrival frame and the temporal relation across frames.

To fully exploit diverse relations among different objects, we propose two instantiations of our graph module: \textbf{visual graph} and \textbf{location graph}. The visual graph is built based on the visual similarity of RoIs to link the same or similar objects and model their relations. The location graph is built on locations/coordinates of RoIs. Spatially overlapped or close objects are connected in the location graph.
The \textbf{streaming} nature of our proposed methods enables the recognition of actions with only a few starting frames. As more frames come in, the accuracy of our model increases steadily. Our graph module is generic and can be combined with any 2D or 3D ConvNet in a plug-and-play fashion.

To demonstrate the effectiveness of our dynamic graph module in improving recognition performance of the backbone network, we conduct experiments on three datasets: Something-Something v1~\cite{goyal2017something}, v2~\cite{mahdisoltani2018fine}, and ActivityNet~\cite{caba2015activitynet}. All datasets consist of videos involving human-object interactions. Videos in Something-Something are short, trimmed, and single-labeled, while videos in ActivityNet are long, untrimmed and multi-labeled. 
Our experimental results support that our graph module can both process streaming videos and help improve the overall performance of existing action recognition pipelines.



\section{Dynamic Graph Modules}
\label{sec:graph}
\head{Definition and Notations.} We denote a video as $V = \{f_1, f_2, ..., f_T\}$ where $f_t$ represents 
the feature map of the $t$-th frame extracted by a 2D ConvNet or the $t$-th feature map extracted by a 3D ConvNet. For each feature map, we keep its top-$N$ region proposals generated by a Region Proposal Network (RPN)~\cite{ren2015faster} and denote the set of proposals as $\bm{B}^t=\{\bm{b}^t_1, \bm{b}^t_2, \dots, \bm{b}^t_N\}$, where the superscript denotes the frame index and the subscript indexes proposals in the current frame.
We represent proposals by their coordinates and extract the associated region feature $\bm{b}^t_n \in \mathbb{R}^{1024}$~\cite{ren2015faster}.
Analogous to the \textit{hidden state} in LSTM, we maintain a \textit{hidden graph} when chronologically processing the video, where we use proposals at $t = 1$ to initialize the hidden graph. We define the \textit{hidden graph} as $\mathcal{G} = (\mathcal{X}, \mathcal{E})$, where $\mathcal{X}=\{\bm{x}_1, \bm{x}_2, \dots, \bm{x}_M\}$ denotes the set of nodes and $\mathcal{E}=\{E(\bm{x}_m, \bm{x}_k)\}$ denotes the set of weighted edges. Here, we allow self-connections within the hidden graph. Each node in the hidden graph has a feature vector and a pair of (virtual) coordinates (top-left, bottom-right). For simplicity, we also use $\bm{x_m} \in \mathbb{R}^{1024}$ to denote the feature of the $m$-th node in the hidden graph and use $(m_{x,1}, m_{y,1}, m_{x,2}, m_{y,2})$ to denote the coordinates of this node.

\begin{figure}[t]
\centering     
\subfigure[The unrolled version of our graph network (backbone ConvNet and RPN are omitted).]{\label{fig:unroll}\includegraphics[width=0.48\linewidth]{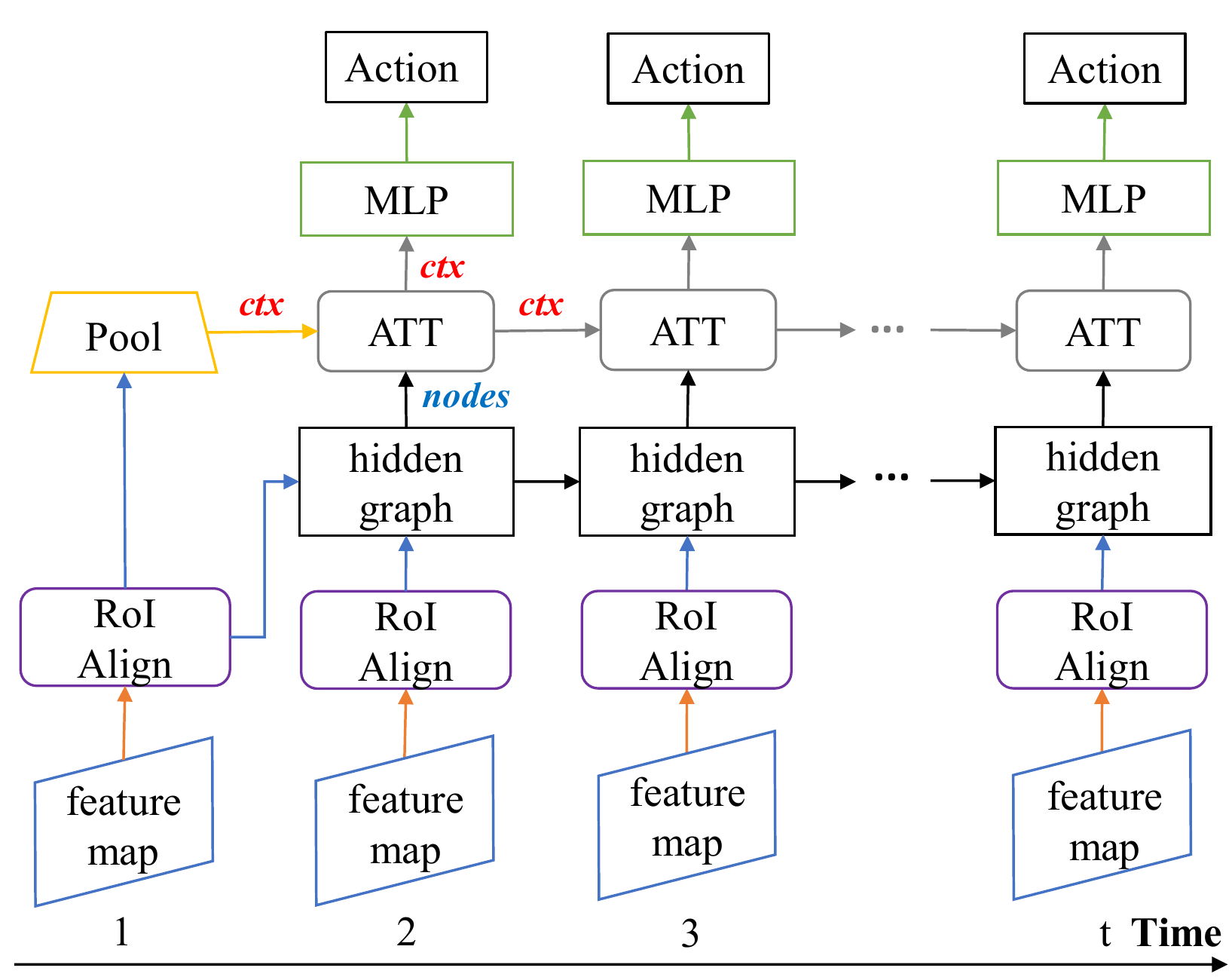}}
\subfigure[The graph building process at each time step $t-1$ and $t$.]{\label{fig:graphbuilding}\includegraphics[width=0.48\linewidth]{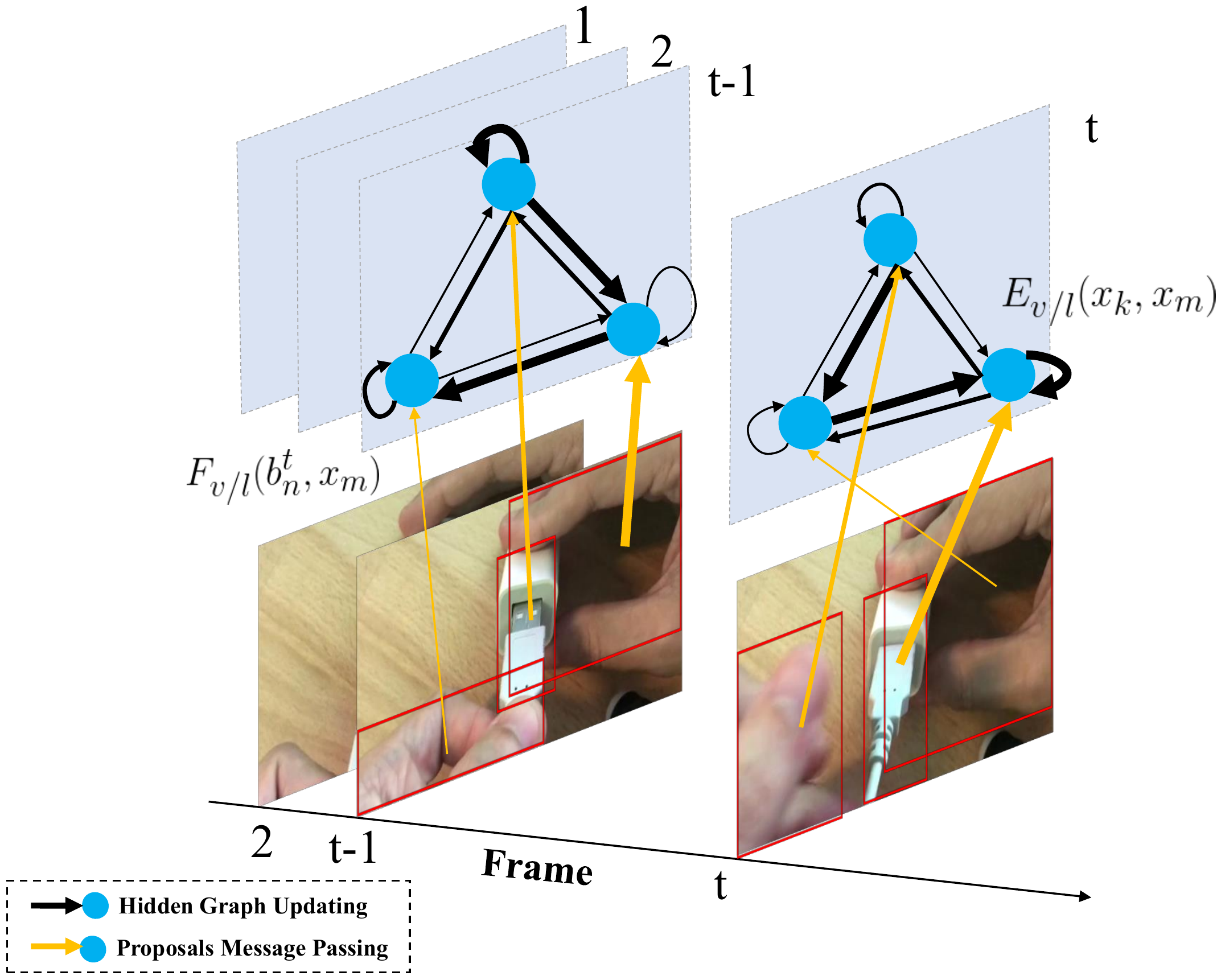}}
\caption{(a).  A ``hidden graph'' is built dynamically in the temporal domain. At each time step, the hidden graph incorporates information from proposals and generates a context vector (denoted as ``\textit{ctx}'' in the figure, more details in Sec.~\ref{attention}) for action recognition. 
(b). At $t-1$ time step, the hidden graph (top row) first incorporates messages from all proposals in the current frame (bottom row) as indicated by yellow arrows; then the hidden graph updates its edges as indicated by black arrows. The width of arrows represents the amount of information that flows along the edges. 
This process iterates in the following time steps.}\vspace{-10pt}
\end{figure}

\head{Graph Module Overview.} In Fig.~\ref{fig:unroll}, we provide an unrolled version of our dynamic graph module where we omit the backbone network and RPN for simplicity. During the initialization, we use max-pooling to summarize all proposals in the first feature map as an initial context vector to warm start our graph module. For each of the following feature maps, proposals are fed into the graph module to update the structure of the hidden graph via an explicit information passing process. We design two types of hidden graphs, visual graph and location graph, based on two different dynamic updating strategies which will be elaborated in Sec.~\ref{visual_graph} and Sec.~\ref{location_graph}. At each time step, the hidden graph contains both visual features and interaction information of different regions accumulated in all previous time steps. We apply a global aggregation function to select a group of the most relevant and discriminative regions to recognize actions. 
More details are provided in Sec.~\ref{attention}.

\subsection{The Visual Graph} 
\label{visual_graph}
Our visual graph aims to link objects with similar appearances/motions and is built based on proposal features. The graph building process is illustrated in Fig.~\ref{fig:graphbuilding}. We use the features of top-$N$ proposals at $t=1$ time step to initialize the features of all nodes in the hidden graph. At time step $t>1$, we measure the pairwise visual similarity between the $N$ proposals in the $t$-th feature map and the $M$ nodes in the hidden graph. The visual similarity is defined as:
\begin{equation}\small
\label{eq:vis_temporal_sim}
\bm{F_v}(\bm{b}^t_n, \bm{x}_m) = h(\bm{b}^t_n)^\top g({\bm{x}_m})
\enspace,
\end{equation}
where $n=1,2,\ldots,N$, $m=1,2,\ldots,M$, and both $h(\cdot)$ and $g(\cdot)$ are linear transformations. We apply $\mathrm{softmax}(\cdot)$ to normalize the weights of edges connecting the $m$-th node in the hidden graph and all proposals in the $t$-th feature map, so that we have:
\begin{equation}\small
\label{eq:vis_temporal_sim_norm}
\bm{F_v}^\prime(\bm{b}^t_n, \bm{x}_m) = \frac{\exp \bm{F_v}(\bm{b}^t_n, \bm{x}_m)}{\sum_{n = 1}^N\exp \bm{F_v}(\bm{b}^t_n, \bm{x}_m)}
\enspace.
\end{equation}
Each node in the hidden graph incorporates information from all $N$ proposals of the $t$-th feature map gated by $\bm{F_v}^\prime(\bm{b}^t_n, \bm{x}_m)$. Therefore, the total amount of inflow information gathered from the $t$-th feature map to node $m$ is:
\begin{equation}\small
\label{eq:vis_temporal_income}
\hat{\bm{x}}_m = \sum_{n=1}^N \bm{F_v}^\prime(\bm{b}^t_n, \bm{x}_m)h(\bm{b}^t_n)
\enspace.
\end{equation}
An intuitive explanation is that each node in the hidden graph looks for the most visually similar proposals and establishes a connection based on the similarity. Subsequently, the node updates its state by absorbing the incoming information:
\begin{equation}\small
\label{eq:vis_temporal_update}
\sigma_v = \textrm{sigmoid}(W_m\bm{x}_m + \hat{W}_m \hat{\bm{x}}_m)
\;, \quad
\bm{x}_m \coloneqq \sigma_v\bm{x}_m + (1 - \sigma_v)\hat{\bm{x}}_m \;, 
\end{equation}
where $\sigma_v$ denotes the gate function controlled by the node state and incoming information, $W_m\in\mathbb{R}^{1024\times1024}$ and $\hat{W}_m\in\mathbb{R}^{1024\times1024}$ are learnable weights. 
If a proposal and a node are more visually similar in the projected space, more information will flow from this proposal to the node. 

After incorporating the information from all $N$ proposals of the $t$-th feature for all nodes, the hidden graph will have an internal update. Notice that the hidden graph is a complete directed graph initially including self-connections. The edge weights are computed as:
\begin{equation}\small
\label{eq:vis_spatial_sim}
\bm{E_v}(\bm{x}_k, \bm{x}_m) = \phi(\bm{x}_k)^\top\phi({\bm{x}_m})
\enspace,
\end{equation}
where $\phi(\cdot)$ is a linear layer with learnable parameters. Eq. \ref{eq:vis_spatial_sim} is similar to Eq. \ref{eq:vis_temporal_sim}, except that both $\bm{x}_m$ and $\bm{x}_k$ are features of nodes in the hidden graph. After the edges of the hidden graph are updated, we propagate information for each node inside the hidden graph using a strategy similar to Eqs.~\ref{eq:vis_temporal_sim_norm}, \ref{eq:vis_temporal_income}, and \ref{eq:vis_temporal_update}. Note that for Eqs. \ref{eq:vis_temporal_sim_norm} and \ref{eq:vis_temporal_income}, we replace $\bm{b}^t_i$ with $\bm{x}_k$, and replace $h(\cdot)$ with $\phi(\cdot)$. Due to the different normalizations, $\bm{E_v}^\prime(\bm{x}_m,\bm{x}_k)$ differs from $\bm{E_v}^\prime(\bm{x}_k,\bm{x}_m)$, hence a directed graph. Moving to the next time step $t + 1$, we repeat the above process. Taking advantage of the iterative processing, our model is capable of processing streaming videos.

\subsection{The Location Graph}
\label{location_graph}

To utilize the displacement of objects to capture spatial relations among proposals, we propose a location graph built upon the coordinates of proposals to link objects that are overlapped or at close positions. 

At time step $t$, the location-based relation between the $n$-th proposal in the $t$-th feature map and the $m$-th node in the hidden graph is defined as:
\begin{equation}\small
\label{eq:loc_temporal_sim}
\bm{F_l}(\bm{b}^t_n, \bm{x}_m) = \bm{\sigma}^t_{n, m}
\enspace,
\end{equation}
where $\bm{\sigma}^t_{n, m}$ represents the value of Intersection-over-Union (IoU) between the $n$-th box in the $t$-th feature map and the $m$-th node in the hidden graph. Similar to~\cite{wang2018videos}, we adopt L-1 norm to normalize weights connecting the $m$-th node in the hidden graph and all proposals in the $t$-th feature map:
\begin{equation}\small
\label{eq:loc_temporal_sim_norm}
\bm{F_l}^\prime(\bm{b}^t_n, \bm{x}_m) = \frac{\bm{F_l}(\bm{b}^t_n, \bm{x}_m)}{\sum_{n = 1}^N\bm{F_l}(\bm{b}^t_n, \bm{x}_m)}
\enspace.
\end{equation}
Analogous to the information passing process in the visual graph, each node in the hidden graph receives messages from all connected proposals from the $t$-th feature map:
\begin{equation}\small
\label{eq:loc_temporal_income}
\hat{\bm{x}}_m = \sum_{n=1}^N \bm{F_l}^\prime(\bm{b}^t_n, \bm{x}_m)p(\bm{b}^t_n) \;, \quad
\bm{x}_m \coloneqq \text{ReLU}(\bm{x}_m + \hat{\bm{x}}_m)
\;,
\end{equation}
where $p(\cdot)$ is a linear transformation. After the information is passed from all proposals to the hidden graph, we update edges in the hidden graph dynamically. We compute IoU between each pair of nodes inside the hidden graph using Eq. \ref{eq:loc_spatial_sim} which is similar to Eq. \ref{eq:loc_temporal_sim}:
\begin{equation}\small
\label{eq:loc_spatial_sim}
\bm{E_l}(\bm{x}_k, \bm{x}_m) = \bm{\sigma}_{k, m}
\enspace,
\end{equation}
where $\bm{x}_k$ and $\bm{x}_m$ are features of nodes in the hidden graph. After the graph is built, messages can be propagated by applying Eqs. \ref{eq:loc_temporal_sim_norm}, \ref{eq:loc_temporal_income}, and \ref{eq:loc_spatial_sim} inside the hidden graph, where we replace $\bm{b}^t_i$ with $\bm{x}_k$, and replace $p(\cdot)$ with another linear transformation $\psi(\cdot)$.

\head{Coordinates updating.} One problem in building the location graph is how to decide the coordinates (``virtual'' bounding box) of each node in the hidden graph.
We propose a \textit{coordinate shifting} strategy to approximate the coordinates of each node in the hidden graph. 

We use the coordinates of the top-$N$ proposals at time step $t=1$ to initialize the coordinates of all nodes in the hidden graph. At time step $t>1$, suppose the top-left and bottom-right coordinates of the $m$-th node in hidden graph are $(m^{t-1}_{x,1}, m^{t-1}_{y,1}, m^{t-1}_{x,2}, m^{t-1}_{y,2})$, and the coordinates of the $n$-th proposal in the $t$-th feature map are $(n^t_{x,1}, n^t_{y,1}, n^t_{x,2}, n^t_{y,2})$. The normalized weight (IoU) between the $m$-th node in the hidden graph and the $n$-th proposal in the $t$-th feature map is $\bm{F_l}^\prime(\bm{b}^t_n, \bm{x}_m)$. The larger the weight, the more information will flow from the $n$-th proposal to the $m$-th node and the coordinates of the $m$-th node will shift more towards the position of the $n$-th proposal. After information passing, the target position of the $m$-th node is the center of the current position and the weighted average positions of all proposals in the $t$-th feature map connected to the $m$-th node. Formally, the coordinate of $m^{t}_{x,1}$ is computed as:
\begin{equation}\small
\label{eq:loc_box_shifting}
m^{t}_{x,1} = \frac{1}{2}(m^{t-1}_{x,1} + \sum_{n=1}^N\bm{F_l}^\prime(\bm{b}^t_n, \bm{x}_m)n^t_{x,1}) \enspace,
\end{equation}
Similarly for $m^{t}_{y,1}$, $m^{t}_{x,2}$ and $m^{t}_{y,2}$ which can be found in the Appendix. Hence, coordinates attached to nodes in the hidden graph will update dynamically according to input proposals at each time step.

\subsection{Attention on Graph}
\label{attention}
At each time step, the hidden graph contains accumulated information from all preceding time steps. The recognition decision is generated based on the state of the hidden graph. We need an aggregation function $\rho$ to gather information from all nodes in the hidden graph. At the same time, such a function should be invariant to permutations of all nodes~\cite{battaglia2018relational}.

Attention mechanism was first proposed in~\cite{bahdanau2014neural} and it takes a weighted average of all candidates based on a query~\cite{kim2017structured}. We add a virtual node to summarize the hidden graph at each time step (see the ``ATT'' block in Fig.~\ref{fig:unroll}). The feature of this virtual node serves two purposes: one is to recognize actions at the current time step, and another is to act as a query (or context) to aggregate information from the hidden graph at the next time step. Specifically, the feature of virtual node at time step $t$ is denoted as $\bm{q}_t$, the feature of $m$-th node in hidden graph at time step $t+1$ is denoted as $\bm{x}^{t+1}_m$, then the feature of virtual node at time step $t+1$, denoted as $\bm{q}_{t+1}$, is computed as:
\begin{align}\small
\label{eq:att_all}
e^{t+1}_m = \tanh{(W_g\bm{q}_t + W_h\bm{x}^{t+1}_m)} \;, \quad \alpha^{t+1}_m = \frac{\exp W_o e^{t+1}_m}{\sum_{m=1}^M \exp W_o e^{t+1}_m} \;, \quad \bm{q}_{t+1} = \sum_{m=1}^M\alpha^{t+1}_m\bm{x}^{t+1}_m \;,
\end{align}
where $W_c$, $W_h$ and $W_o$ are learnable weights. Note that the initial feature of the virtual node is the max-pooling of all proposals in the first feature map. Once the feature of the virtual node is generated, we can forward the feature into a multi-layer perceptron to recognition actions.

\section{Full Model for Action Recognition}
\label{sec:full}

In this section, we 
introduce two versions of our full models: streaming version and static version. 
The streaming version can process streaming videos while the static version incorporates the global video feature and achieves better overall performance.

\head{Streaming Version.} Given a video clip (around 5 seconds), our model first randomly samples 32 frames. The sampled frames are fed into a backbone network. In our case, we apply a 3D ConvNet~\cite{carreira2017quo}. The output of the backbone is a sequence of 3D feature maps with the shape of $T\times C\times H\times W$ .
We apply a region proposal network (RPN)~\cite{ren2015faster} to extract proposals for each sampled frame. With the proposed bounding boxes, we conduct RoIAlign~\cite{he2017mask} on the sequence of feature maps. We build our graph module dynamically upon a sequence of RoI proposals from the feature maps. We maintain a ``hidden graph'' which evolves along the temporal dimension and generates a recognition result at each time step. 

\head{Static Version.} To achieve better recognition accuracy, it is beneficial to utilize all information contained in a video. We provide a static version of our model in which 
we sample frames from an entire video and input all sampled frames into both the backbone 3D ConvNet and a RPN. We average pool the features produced by the 3D ConvNet from $T\times C\times H\times W$ to $C \times 1$, denoted as $\bm{f}$. Different from the streaming version where we only use the graph module feature $\bm{q}_t$ at each time step, here we fuse both graph module features and 3D ConvNet features by concatenating them to recognize actions. More details about the fusion layers are in the Appendix.


\section{Experiment}
\label{sec:exp}

\subsection{Datasets, Metrics, and Implementation Details}

\head{Datasets.} We evaluate our dynamic graph module on datasets: Something-Something v1 and v2~\cite{goyal2017something, mahdisoltani2018fine} and ActivityNet~\cite{caba2015activitynet}.
%
Something-Something v1~\cite{goyal2017something} contains more than 100K short videos and v2~\cite{mahdisoltani2018fine} contains around 220K videos. The average video duration is about 3 to 6 seconds. There are 174 total action classes and each video corresponds to exactly one action. For both v1 and v2 datasets, we follow the official split to train and test our model.
%
ActivityNet contains 10K videos for training, enclosing 15K activity instances from 200 activity classes. The validation set contains 5K videos and 7K activity instances.
We also follow the official split to train and test our model.


\head{Metrics.} Since all videos in Something-Something dataset are single-labeled, we adopt recognition accuracy (top-k) as our evaluation metrics. In ActivityNet dataset, mean average precision (mAP) is also used for prediction evaluation as some videos have multiple labels.

\head{Compared methods.} To verify that our dynamic graph module is capable of modeling interactions between objects, we design a baseline LSTM model where we feed in the mean-pooled top-$N$ region features at each time step.
We compare our streaming model with this baseline, along with a state-of-the-art method~\cite{zolfaghari2018eco}.
We also compare our full static model with competitive existing works~\cite{zhou2018temporal, lee2018motion, wang2018videos, zolfaghari2018eco, lin2018temporal, shao2018high}. 

\head{Region Proposal and Feature.} For each input frame, we propose RoI proposals using RPN with ResNet-50 pre-trained on Microsoft COCO. We project proposal coordinates from the input frames back to the feature maps generated by the penultimate convolutional block of 3D backbone. Since 32 input frames are reduced to 8 feature maps in the temporal domain, we select 8 input frames (i.e., 1-th, 5-th, 9-th, ...) to match the 8 feature maps
. We apply RoIAlign~\cite{he2017mask} with the same configuration in~\cite{wang2018videos} to extract features for each proposal.

\head{Training.} For the backbone network, we follow the frame sampling and training strategy in~\cite{wang2018videos}. Then for our full model, we fix the backbone 3D ConvNet and only train other parts, \eg, our graph module, fusion layers and classification layer. We adopt the same learning strategy as the fine-tuning of the backbone. More details are in the Appendix.

\head{Inference.} 
For Something-Something dataset, we uniformly sample 32 frames from the entire video and rescale them with the shorter side to 256. Then we center crop each frame to $224\times 224$. For ActivityNet dataset, we segment each video into 5s long clips without overlapping and uniformly sample 32 frames from each clip. We adopt top-k pooling to average scores of all clips as the video-level score. 


\subsection{Results of Streaming Model}

\begin{wrapfigure}{r}{0.5\textwidth}
\centering
\includegraphics[width=0.5\textwidth]{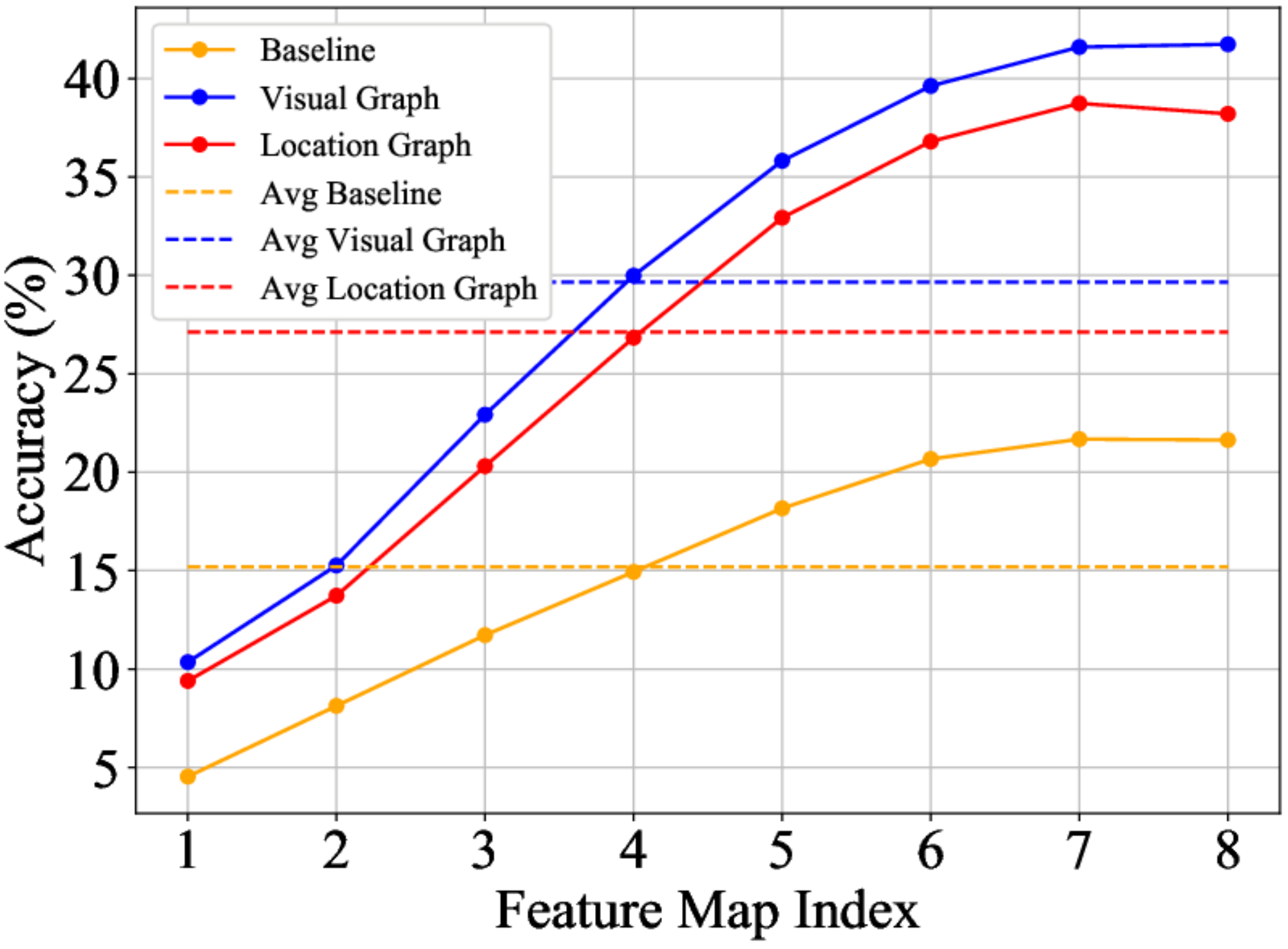}
\caption{Top-1 accuracy on Something-Something v1 validation set for each feature map. ``Avg'' means the average accuracy of the total eight feature maps.}
\label{fig:sth_v1_frame}
\vspace{-10pt}
\end{wrapfigure}

Videos in the Something-Something dataset usually contain two to three objects including humans. We keep the top 20 region proposals for each frame and fix the number of nodes in the hidden graph to 5. We plot the top-1 accuracy in Fig.~\ref{fig:sth_v1_frame}.

The accuracy of the baseline model is significantly lower than any of our graph modules, indicating that feeding the average pooling over proposals into an LSTM fails to capture interactions between objects. One possible explanation is that the average pooling operation discards the spatial relations contained in proposals. The only temporal relation modeled by LSTM is insufficient to capture interactions. On the contrary, as our graph module maintains a graphical structure to keep both spatial and temporal relations among proposals, it has the capability to model the complex interactions among objects.

Between the two graph modules, we notice that the visual graph outperforms the location graph. That is possibly because the visual graph contains more parameters than the location graph, which gives the visual graph more powerful modeling ability. Though the location graph performs inferior than the visual graph, it still achieves more than 38.5\% top-1 accuracy. It is reasonable to conclude that the graph module structure intrinsically has the ability to model interactions regardless of any specific instantiation.

%
%
%

%
%
\begin{table}[t]\footnotesize
\centering
\begin{minipage}[t]{0.55\linewidth}
\begin{center}
\begin{tabular}{l|cc|cc}
\toprule
Model                 & \multicolumn{2}{c|}{Someth. v1} & \multicolumn{2}{c}{Someth. v2} \\ \hline
\multirow{2}{*}{Ours} & Visual     & Location   & Visual     & Location    \\
                      & 41.7              & 38.2                & 54.0              & 50.5               \\ \hline
ECO Lite~\cite{zolfaghari2018eco}              & \multicolumn{2}{c|}{41.3}               & \multicolumn{2}{c}{-}  \\                \bottomrule
\end{tabular}
\end{center}
\caption{Top-1 accuracy of the last feature map on Someth. validation set. (``Visual'' and ``Location'' refer to ``visual graph'' and ``location graph''.)}
\label{tab:streammodel_sth}
\end{minipage}\hfill
\begin{minipage}[t]{0.4\linewidth}
\begin{center}
\begin{tabular}{l|ccc}
\toprule
\multicolumn{1}{c|}{model} &  mAP & top-1 & top-3  \\ \hline
Backbone                 & 70.2 &  69.2 &   83.5   \\
Visual Graph               & 71.5 &  \textbf{70.3} &   84.5   \\
Location Graph             & \textbf{71.8} &  \textbf{70.3} &   84.8   \\
\bottomrule
\end{tabular}
\end{center}
\caption{Results of the static version model on ActivityNet dataset. With RGB frame inputs only.}
\label{tab:staticmodel_act}
\end{minipage}
\vspace{-10pt}
\end{table}

The accuracy of the two graph modules increases steadily as the number of frames increases
and plateaus at 7-th feature map.
It demonstrates that our graph module has the ability to recognize actions in streaming videos, even if only parts of frames are forwarded into the module. We also report the accuracy of the last feature map in Table~\ref{tab:streammodel_sth}. On Something-Something v1 dataset, our visual graph performs slightly better than ~\cite{zolfaghari2018eco} which is a recent state-of-the-art streaming method. Distinct from~\cite{zolfaghari2018eco}, our model explicitly focuses on modeling object interactions.  
The location graph does not perform as competitive as the visual graph. However, note that the location graph has fewer parameters as illustrated in Sec.~\ref{location_graph}.

\begin{table}[]\footnotesize
\begin{center}
{%
\begin{tabular}{llccccccc}
\toprule
                  & & \multicolumn{3}{c}{Something v1}                                & \multicolumn{4}{c}{Something v2}                              \\
                  & \multicolumn{1}{c|}{} & \multicolumn{2}{c|}{val}           & \multicolumn{1}{c|}{test}  & \multicolumn{2}{c|}{val}           & \multicolumn{2}{c}{test} \\
                  & \multicolumn{1}{c|}{Modality} & top-1 & \multicolumn{1}{c|}{top-5} & \multicolumn{1}{c|}{top-1} & top-1 & \multicolumn{1}{c|}{top-5} & top-1       & top-5      \\ \hline
2-Stream TRN~\cite{zhou2018temporal}    & \multicolumn{1}{c|}{Flow+RGB}  & 42.0  & \multicolumn{1}{c|}{-}     & \multicolumn{1}{c|}{40.7}  & 55.5  & \multicolumn{1}{c|}{83.1}  & 56.2         & 83.2        \\
MFNet-C101~\cite{lee2018motion}     & \multicolumn{1}{c|}{RGB only}   & 43.9  & \multicolumn{1}{c|}{73.1}  & \multicolumn{1}{c|}{37.5}   & -     & \multicolumn{1}{c|}{-}     & -         & -        \\
Space-Time Graphs~\cite{wang2018videos} & \multicolumn{1}{c|}{RGB only} & 46.1  & \multicolumn{1}{c|}{\textbf{76.8}}  & \multicolumn{1}{c|}{\textbf{45.0}}   & -     & \multicolumn{1}{c|}{-}     & -         & -        \\
ECO$_{En}Lite$~\cite{zolfaghari2018eco} & \multicolumn{1}{c|}{RGB only} & 46.4  & \multicolumn{1}{c|}{-}  & \multicolumn{1}{c|}{42.3}   & -     & \multicolumn{1}{c|}{-}     & -         & -        \\
ECO$_{En}Lite$~\cite{zolfaghari2018eco} & \multicolumn{1}{c|}{Flow+RGB} & \textbf{49.5}  & \multicolumn{1}{c|}{-}  & \multicolumn{1}{c|}{43.9}   & -     & \multicolumn{1}{c|}{-}     & -         & -        \\
TSM$_{16F}$~\cite{lin2018temporal} & \multicolumn{1}{c|}{RGB only} & 44.8  & \multicolumn{1}{c|}{74.5}  & \multicolumn{1}{c|}{-}   & 58.7     & \multicolumn{1}{c|}{84.8}     & \textbf{59.9}         & \textbf{85.9}        \\
LEGO~\cite{shao2018high} & \multicolumn{1}{c|}{RGB only} & 45.9  & \multicolumn{1}{c|}{-}  & \multicolumn{1}{c|}{-}   & 59.6     & \multicolumn{1}{c|}{-}     & -         & -        \\
\hline
Backbone         & \multicolumn{1}{c|}{RGB only} & 46.0  & \multicolumn{1}{c|}{76.1}  & \multicolumn{1}{c|}{-}   & 59.7  & \multicolumn{1}{c|}{86.4}  & -         & -        \\
Visual Graph     & \multicolumn{1}{c|}{RGB only} & \textbf{47.1}  & \multicolumn{1}{c|}{76.2}  & \multicolumn{1}{c|}{-}   & \textbf{61.4}  & \multicolumn{1}{c|}{\textbf{86.8}}  & -         & -        \\
Location Graph  & \multicolumn{1}{c|}{RGB only}  & \textbf{47.1}  & \multicolumn{1}{c|}{\textbf{76.3}}  & \multicolumn{1}{c|}{\textbf{44.5}}   & \textbf{61.4}  & \multicolumn{1}{c|}{\textbf{86.8}}  & \textbf{59.7}         & \textbf{86.1}  \\
\bottomrule
\end{tabular}
}
\end{center}
\caption{Comparing performance of the static version model on Something-Something v1 \& v2 datasets with state-of-the-art methods. The ``test'' columns are leaderboard results. Note that we only use RGB modality and relatively simple preprocessing steps. The top two scores of each metric are highlighted. (``-'' means there is no publicly available evaluation scores released by the authors.)}
\label{tab:staticmodel_sth}\vspace{-10pt}
\end{table}

\subsection{Results of Static Model}
\noindent\head{Something-Something Dataset.} We compare our static version model with some recent works~\cite{zhou2018temporal, lee2018motion, wang2018videos, zolfaghari2018eco, lin2018temporal, shao2018high} shown in Table~\ref{tab:staticmodel_sth}. For Something-Something v1 validation set, the backbone 3D ConvNet has achieved 46.0\% in top-1 accuracy and 76.1\% in top-5 accuracy. By adding our two types of dynamic graph modules to the backbone, the performance improves an absolute 1.1\%. For Something-Something v2 validation set, the backbone 3D ConvNet has achieved 59.7\% in top-1 and 86.4\% in top-5 accuracy. Our graph module still boosts the performance of the backbone by an absolute 1.7\% for top-1 accuracy. We also report our results on the leaderboard (results shown in the ``test'' column). Without bells and whistles (\eg, flow inputs and ensembling), our model achieves competitive results.


\noindent\head{ActivityNet Dataset.} We also evaluate our static version model on ActivityNet dataset and report the result in Table~\ref{tab:staticmodel_act}. Different from trimmed and shorter videos in Something-Something dataset, videos in ActivityNet are untrimmed and longer, and some contain multiple actions. The backbone 3D ConvNet has achieved 69.2\% top-1 accuracy and the mAP is 70.2\%\footnote{The 3D backbone network is our own implementation.}. Note that compared with the state-of-the-art performance~\cite{wang2016temporal,Xiong2016CUHKE}, we only apply random rescale and random horizontal flip to RGB images without any other complicated data augmentation. We also do not use audio modality, optical-flow features or ensembles, etc. Both types of dynamic graph modules bring around 1.5\% improvement in mAP compared to the backbone. The result demonstrates our module's capability on long-term action recognition in untrimmed videos. 
As our model is trained on trimmed action instances level by sampling a fixed number of frames but tested on whole videos, we can draw a conclusion that 
our proposed graph module is capable of recognizing actions in both single-labeled trimmed videos and multi-labeled untrimmed videos. 
\section{Related Works}
\label{sec:related}


\head{Video action recognition with deep learning.} 
Many works have applied convolutional networks to tackle video action recognition problems~\cite{karpathy2014large, simonyan2014two, tran2014c3d, tran2017convnet, carreira2017quo, wang2018non, wang2016temporal}. Karpathy \etal~\cite{karpathy2014large} explored various approaches of fusing RGB frames in temporal domain. Simonyan \etal~\cite{simonyan2014two} devised a two-stream model to fuse RGB features and optical flow features. Tran \etal~\cite{tran2014c3d} applied a 3D kernel to convolve a sequence of frames in spatiotemporal domain. 
\cite{carreira2017quo} proposed inflated 3D convolutional networks (I3D) which utilize parameters in 2D ConvNets pre-trained on ImageNet~\cite{krizhevsky2012imagenet}. \cite{wang2016temporal} proposed the temporal segment network (TSN) 
which sparsely sampled frames. Zhou \etal~\cite{zhou2018temporal} showed that the order of frames is crucial for correct recognition. 
Zolfaghari \etal~\cite{zolfaghari2018eco} proposed an online video understanding system combining 2D ConvNets and 3D ConvNets. 
ConvNets is also one of the components in our model. However, ConvNets lack the power to model explicit object interactions, which is the problem the proposed module aims to solve.

\head{Relational model / Graph neural networks.} Another line of work in action recognition is focusing on modeling object relationships. 
Ma \etal~\cite{ma2018attend} utilized an LSTM to model object interactions but lost spatial information. Wang \etal~\cite{wang2018non} added a non-local layer to 3D ConvNets to capture relations among different positions in feature maps. However, two distant positions are generally likely to be irrelevant.
Some works apply graph neural network (GNN) to model object relations. 
\cite{chen2018graph} projected pixels to graph space and then projected back to build relations among different regions, but it cannot guarantee each region corresponds to (a part of) an object. 
The most similar work to ours is~\cite{wang2018videos}, where a video is represented as a global space-time graph of object regions. 
We propose to use a dynamic hidden graph to process sequential video input, in the form of object region proposals, which takes advantage of both relational modeling and sequential modeling~\cite{donahue2015long}.


\section{Conclusion}
\label{sec:conclusion}

\newcommand{\la}[1]{\renewcommand{\arraystretch}{#1}}
We propose a novel dynamic graph module with two instantiations, visual graph and location graph, to model object-object interactions in video activities. By considering object relations in spatial and temporal domains simultaneously, the proposed graph module can capture interactions among objects explicitly in streaming video settings, which differs our work from existing methods. We will extend our graph module to more sequential modeling fields, \eg. video prediction, in the future.

\paragraph{Acknowledgement.} 
H. Huang, W. Zhang, and C. Xu are supported by NSF IIS 1813709, IIS 1741472, and CHE 1764415. L. Zhou and J. J. Corso are supported by DARPA FA8750-17-2-0125, NSF IIS 1522904 and NIST 60NANB17D191. This article solely reflects the opinions and conclusions of its authors but not the DARPA, NSF, or NIST.

{\small
\bibliography{egbib}
}

\newpage
\appendix
\section{Appendix}
This Appendix provides additional algorithm formulas, network structures and implementation details.

\subsection{Coordinates updating}
\label{sec:coords}
At time step $t>1$, suppose the top-left and bottom-right coordinates of the $m$-th node in hidden graph are $(m^{t-1}_{x,1}, m^{t-1}_{y,1}, m^{t-1}_{x,2}, m^{t-1}_{y,2})$, and the coordinates of the $n$-th proposal in the $t$-th feature map are $(n^t_{x,1}, n^t_{y,1}, n^t_{x,2}, n^t_{y,2})$. The normalized weight (IoU) between the $m$-th node in the hidden graph and the $n$-th proposal in the $t$-th feature map is $\bm{F_l}^\prime(\bm{b}^t_n, \bm{x}_m)$. The coordinate of $m^{t}_{x,1}$ is computed as:
\begin{equation}
\label{eq:loc_box_shifting}
\left\{
    \begin{array}{ll}
        m^{t}_{x,1} = \frac{1}{2}(m^{t-1}_{x,1} + \sum_{n=1}^N\bm{F_l}^\prime(\bm{b}^t_n, \bm{x}_m)n^t_{x,1}) \enspace,\\
        m^{t}_{y,1} = \frac{1}{2}(m^{t-1}_{y,1} + \sum_{n=1}^N\bm{F_l}^\prime(\bm{b}^t_n, \bm{x}_m)n^t_{y,1}) \enspace,\\
        m^{t}_{x,2} = \frac{1}{2}(m^{t-1}_{x,2} + \sum_{n=1}^N\bm{F_l}^\prime(\bm{b}^t_n, \bm{x}_m)n^t_{x,2}) \enspace,\\
        m^{t}_{y,2} = \frac{1}{2}(m^{t-1}_{y,2} + \sum_{n=1}^N\bm{F_l}^\prime(\bm{b}^t_i, \bm{x}_m)n^t_{y,2}) \enspace.\\
    \end{array}
\right.
\end{equation}
%
\subsection{The Structure of Fusion Layers}
\label{sec:fusion}
\begin{figure}[ht]
\centering
\includegraphics[width=0.8\linewidth]{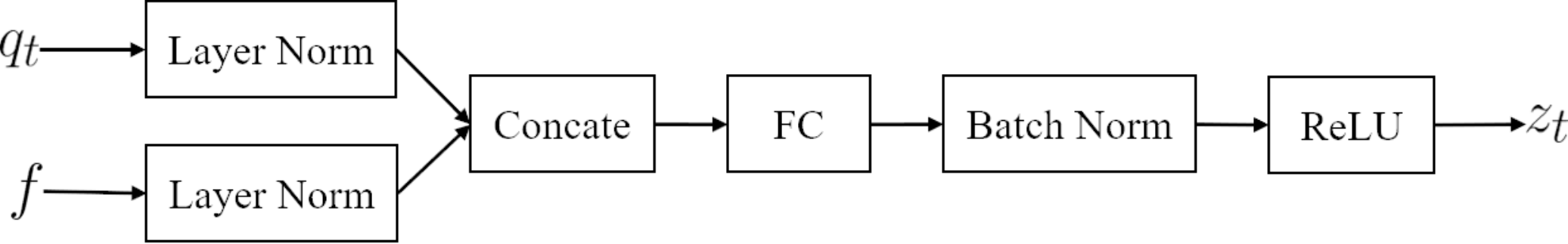}
\caption{Fusion layers to fuse the graph module feature and 3D ConvNet feature at time step $t$.}
\label{fig:fuse}
\end{figure}

The average-pooled feature produced by the 3D ConvNet is denoted as $\bm{f} \in \mathbb{R}^{C\times 1}$ where $C=2048$. The graph module feature $\bm{q}_t \in \mathbb{R}^{C^\prime \times 1}$ where $C^\prime = 1024$. We fuse both graph module feature and 3D ConvNet feature to recognize actions. The fusion layers are illustrated in Fig.~\ref{fig:fuse}. We keep the size of the fused feature $\bm{z}_t$ to $C^\prime \times 1$ and forward this feature into a multi-layer perceptron to get the final recognition results.

\subsection{Implementation Details}
\label{sec:details}
We first train our backbone 3D model~\cite{carreira2017quo, wang2018non} on Kinetics dataset and then fine-tune it on the target datasets. For Something-Something dataset, we randomly sample 32 frames from each video. For ActivityNet dataset, as the video length is much longer, we first segment each activity instance into several clips (around 5 seconds) with the overlap rate fixed to 20\%. The sampled frames are used to train our backbone 3D model. Following \cite{wang2018videos}, sampled frames are randomly scaled with shorter side resized to a random integer number in [256, 320]. Then we randomly crop out an area of $224 \times 224$ and randomly flip frames horizontally before forwarding them to the backbone model. The Dropout \cite{hinton2012improving} before the classification layer in backbone model is set to 0.5. We train our backbone model with a batch size of 24. We set the initial learning rate to 0.00125. We apply stochastic gradient descent (SGD) optimizer and set momentum to 0.9 and weight decay to 0.0001. We adopt cross-entropy loss during our training. We adopt cross-entropy loss during our training.

Next, we describe how we train our streaming dynamic graph module. For each input frame, we  propose RoI proposals using RPN~\cite{ren2015faster} with ResNet-50 pre-trained on Microsoft COCO. For Something-Something dataset, we keep the top 20 proposals each frame and set the number of nodes in hidden graph to be 5. For ActivityNet dataset, as video scenes are more complex and contain more objects, we keep the top 40 proposals and increase the number of graph nodes to 10. We fix the backbone 3D ConvNet and only train our graph module, fusion layers and classification layer. We adopt the same learning strategy as the fine-tuning of the backbone.

For the static model, we first train the streaming model following the strategy above for 3 epochs as a warm-up. Then we concatenate the graph module feature with the backbone feature using the fusion layers described in Sec.~\ref{sec:fusion}. At the same time, we reduce the learning rate by a factor of 10. The parameters of the backbone remain fixed during training.


\end{document}